\documentclass[letterpaper]{article}
\usepackage{iccc}

\usepackage{tabularx}
\usepackage{booktabs}
\usepackage{multirow}
\usepackage{amsmath}
\usepackage{times}
\usepackage{helvet}
\usepackage{courier}
\usepackage{listings}
\usepackage{hyperref}
\usepackage{longtable}
\usepackage{stfloats}
\usepackage{tikz}
\usetikzlibrary{arrows.meta,positioning,fit,backgrounds,calc}

\makeatletter
\renewcommand\paragraph{\@startsection{paragraph}{4}{\z@}%
  {0.5ex \@plus 0.5ex \@minus 0.2ex}%
  {-1em}%
  {\normalfont\normalsize\bfseries}}
\makeatother

\title{Grounding Machine Creativity in Game Design Knowledge Representations:\\
Empirical Probing of LLM-Based Executable Synthesis of Goal Playable Patterns under Structural Constraints}
\author{Hugh Xuechen Liu, Kıvanç Tatar\\
Chalmers University of Technology and University of Gothenburg \\
\texttt{xuechen@chalmers.se, tatar@chalmers.se}
}
\begin{document}
\maketitle

\begin{abstract}
Creatively translating complex gameplay ideas into executable artifacts (e.g., games as Unity projects and code)  remains a central challenge in computational game creativity. Gameplay design patterns provide a structured representation for describing gameplay phenomena, enabling designers to decompose high-level ideas into entities, constraints, and rule-driven dynamics. Among them, \emph{goal patterns} formalize common player--objective relationships. Goal Playable Concepts (GPCs) operationalize these abstractions as playable Unity engine implementations, supporting experiential exploration and compositional gameplay design.

We frame scalable playable pattern realization as a problem of \emph{constrained executable creative synthesis}: generated artifacts must satisfy Unity’s syntactic and architectural requirements while preserving the semantic gameplay meanings encoded in goal patterns. This dual constraint limits scalability. Therefore, we investigate whether contemporary large language models (LLMs) can perform such synthesis under engine-level structural constraints and generate Unity code (as games) structured and conditioned by goal playable patterns

Using 26 goal pattern instantiations, we compare a direct generation baseline (natural language $\rightarrow$ C\# $\rightarrow$ Unity) with pipelines conditioned on a human-authored Unity-specific intermediate representation (IR), across three IR configurations and two open-source models (DeepSeek-Coder-V2-Lite-Instruct and Qwen2.5-Coder-7B-Instruct). Compilation success is evaluated via automated Unity replay. We propose grounding and hygiene failure modes, identifying structural and project-level grounding as primary bottlenecks.
\end{abstract}

\section{Introduction}
\label{sec:introduction}
Computational Creativity (CC) investigates how computational systems generate novel, coherent, and valuable artifacts~\cite{boden2004creative,colton2011computational}. In computational game creativity---CC ``within and for'' digital games~\cite{liapis2014computational}---an additional constraint is unavoidable: artifacts must be executable and playable. We frame this challenge as \emph{creative realization}~\cite{costikyan2002have}: given structured game design knowledge, can a computational system reliably instantiate it as operational and playable content? Executable viability is not merely an engineering gate but a necessary condition for artifact existence under real engine constraints. A second motivation is scale: human-authored playable concept implementations~\cite{kultima2020designing,lyu2023goal} are rich but do not scale; LLM-driven pipelines that reliably instantiate design knowledge as executable artifacts open pathways to large-scale exploration of gameplay design spaces.

We adopt a \emph{representation-first} perspective~\cite{shaker2016procedural,togelius2011search,liapis2016mixed}: rather than treating LLM generation as direct text-to-code synthesis, we ask whether explicit design knowledge encoded as a structured intermediate representation (IR) can scaffold reliable creative realization. Gameplay design patterns~\cite{bjork2005patterns} provide a structured representation for recurring interaction structures; among these, \emph{goal patterns} describe how player objectives are constituted through entities, constraints, and rule-governed interactions. Instantiating a goal pattern in Unity requires coherent object configuration, correct component attachment, and runtime wiring---making goal-pattern realization a stringent testbed for creative realization under real engine constraints.

Using 26 goal pattern reference implementations~\cite{lyu2023goal,kultima2020designing}, we compare a direct generation baseline against pipelines conditioned on IR v0.2-runtime-evidence across three configurations (\texttt{free}, \texttt{min}, \texttt{full}) and two open-source models: DeepSeek-Coder-V2-Lite-Instruct~\cite{dai2024deepseekmoe} and Qwen2.5-Coder-7B-Instruct~\cite{hui2024qwen2,qwen2}. Every generated artifact failed to compile; rather than treating this as a negative result, we analyze the structured distribution of failures as empirical evidence of where grounding breaks down.

\paragraph{Contributions.}
We contribute: \textbf{(1) Creative realization as a CC problem framing}---situating executable pattern instantiation within computational creativity and establishing compile-grounded viability as a necessary condition for artifact existence; \textbf{(2) An execution-grounded evaluation pipeline}---an end-to-end workflow from HPC inference to Unity batch replay and per-seed metric aggregation; \textbf{(3) A Unity-specific IR for knowledge injection}---IR v0.2-runtime-evidence encoding project-level structural conventions and goal-pattern semantic knowledge, evaluated under three conditioning configurations (\texttt{free}, \texttt{min}, \texttt{full}); \textbf{(4) A structured failure taxonomy}---empirical analysis of grounding and hygiene failure modes, revealing project-level and engine-level grounding as primary bottlenecks for knowledge-conditioned LLM generation in Unity.

\paragraph{Note on model scope.}
The present study uses two open-source models (DeepSeek-Coder-V2-Lite-Instruct; Qwen2.5-Coder-7B-Instruct) chosen for HPC batch inference feasibility and open weights. We acknowledge that larger models may exhibit different failure profiles.
Follow-up work with a redesigned IR and Qwen3-Coder-30B-A3B is ongoing, motivated by the hypothesis that model capacity and IR design are separable variables---a question the present taxonomy is designed to diagnose.

\section{Background}

\subsection{Computational Creativity and Constrained Generative Synthesis}
Computational creativity research concerns systems that generate novel, valuable, and surprising artifacts~\cite{boden2004creative,ventura2016beyond,colton2012computational}. Boden's framework distinguishes combinatorial, exploratory, and transformational creativity~\cite{boden2004creative}; exploratory creativity, most relevant here, involves traversing a structured conceptual space to produce artifacts novel within its boundaries. In generative systems, constraints serve a dual role: bounding the conceptual space where creativity can take place.~\cite{wiggins2006preliminary,wiggins2006searching}. This is particularly acute in \emph{executable creative synthesis}, where artifacts must satisfy both semantic design intentions and syntactic execution requirements---unlike open-ended generation, executable artifacts admit objective validity criteria: they either run or they do not.

Co-creative systems distribute creative agency across human and machine contributors~\cite{liapis2016mixed,deterding2017mixed}, with each contributing what the other cannot efficiently provide. Prior work demonstrates that human-authored constraints can substantially improve coherence of machine-generated content~\cite{liapis2016mixed}. Our work instantiates this model: human authors contribute domain knowledge as a structured IR while the model contributes generative breadth; where this boundary breaks down is a central empirical question. Prior computational game creativity work has approached generation through procedural content generation (PCG)~\cite{shaker2016procedural} or LLM-based content generation~\cite{todd2023level,chen2023gamegpt,sudhakaran2023mariogpt}; our work differs in targeting instantiation of a design concept as a complete executable artifact, foregrounding grounding as central to creative synthesis.

\subsection{LLMs for Code Generation and Executable Artifact Synthesis}
Large language models have demonstrated substantial capability from function-level synthesis~\cite{chen2021evaluating} to repository-level editing~\cite{jimenez2023swe,jiang2026survey}, extended to structured generation conforming to domain-specific schemas~\cite{poesia2022synchromesh} and creative domains balancing novelty with formal constraints~\cite{yuan2022wordcraft}. However, LLM-based code generation exhibits characteristic failure modes for complex, architecture-dependent artifacts: models generate locally plausible code that fails at integration---syntactically correct components that are architecturally incompatible within the target project~\cite{jimenez2023swe,chen2023gamegpt}. In game engine contexts this is compounded by engine-specific conventions underrepresented in pretraining corpora: Unity's component model, scene graph architecture, and MonoBehaviour lifecycle impose constraints that differ substantially from general-purpose programming patterns.

Intermediate representations (IR) address this complexity by decomposing generation into semantically meaningful intermediate steps with more tractable targets~\cite{yin2017syntactic,austin2021program}, particularly effective when the IR externalizes stable domain knowledge and reduces the burden on the model to rediscover structural conventions from context alone~\cite{dohan2022language,gao2023pal}.

\subsection{Goal Playable Concepts as Coupled Game Design Representation}
Gameplay design patterns describe recurring interaction structures for analysis and reuse in game design~\cite{bjork2005patterns}. As intermediate-level design knowledge~\cite{hook2012strong}, they occupy a position between abstract theories and concrete instances, encoding interaction relations rather than surface aesthetics. Because gameplay design constitutes second-order design~\cite{tekinbas2003rules}---designers control rules rather than experience directly---pattern descriptions are inherently abstract and resistant to direct operationalization. Patterns constitute a \emph{design language}~\cite{rheinfrank1996design} with densely interconnected instantiation and modulation relations~\cite{bjork2005patterns}, meaning no single pattern can be instantiated without invoking others and imposing non-trivial compositional requirements on any generative system.

Among these, \emph{goal patterns} formalize player--objective relationships focusing on imperative interaction-level goals~\cite{bjork2005patterns,debus2020typology}---a bounded, structured design space amenable to both exploratory and combinatorial creative operations in Boden's sense~\cite{boden2004creative}. Goal Playable Concepts (GPCs)~\cite{lyu2023goal} \footnote{\url{https://gameplaydesignpatterns.itch.io/}} operationalize goal patterns as playable Unity implementations, coupling textual descriptions with interactive instantiations. Neither format is sufficient alone: patterns lack interactivity while playable concepts cannot encode abstract relational structure; their coupling is structurally necessary as each serves as contextual scaffolding for interpreting the other~\cite{rheinfrank1996design,kultima2020designing,consalvo2017paratexts}. From a data perspective, GPCs expose design knowledge across three modalities: natural language descriptions, graph-structured relational knowledge, and executable Unity code---providing a grounded source from which to derive an IR encoding both structural conventions and semantic gameplay intentions.

\subsection{Grounding LLM Generation in Structured Domain Knowledge}
Grounding constrains model outputs to conform to external knowledge structures, reducing the gap between fluent generation and valid artifacts~\cite{ji2023survey}. Staged pipelines with IRs improve conformance by externalizing stable domain knowledge~\cite{dohan2022language,gao2023pal,chen2021evaluating,jiang2026survey}. In game engine contexts, this is compounded by engine-specific conventions underrepresented in pretraining corpora---Unity's component model, MonoBehaviour lifecycle, and scene graph architecture impose constraints that differ substantially from general-purpose programming patterns.

\section{Problem Setting and Method}

\subsection{Problem Setting}
We consider 26 Unity scene implementations, each a reference instantiation of a distinct goal pattern from the Goal Playable Concepts collection, co-existing in an authored Unity project containing prefab assets, MonoBehaviour scripts, and scene-level configuration. Generation targets a new Unity Editor script instantiating the scene for the target pattern, performed on top of this existing layer rather than from scratch. Evaluation measures compilation viability under real Unity engine constraints; semantic fidelity evaluation remains future work.

\subsection{Pipeline Overview}
We compare two pipelines across four configurations (Figure~\ref{fig:pipeline}). \textbf{No-schema baseline}: the model receives the goal pattern as a natural language markdown document and generates a Unity Editor C\# script directly. \textbf{IR-conditioned pipeline}: generation proceeds in two steps---first generating an IR JSON from the pattern description, then translating it to C\#---under three conditioning levels: \texttt{free} (no schema constraints), \texttt{min} (minimal field skeleton), and \texttt{full} (complete IR v0.2-runtime-evidence schema with four hard referential-integrity constraints). Prompt templates are provided verbatim in \nameref{sec:appendix-prompt-template}; Table~\ref{tab:prompts} summarises the key differences across configurations.

\begin{figure}[!t]
\centering
\resizebox{\linewidth}{!}{%
\begin{tikzpicture}[
  node distance=2mm and 3mm,
  box/.style={draw, rounded corners=1pt, align=center, minimum height=5mm, minimum width=11mm, inner sep=1pt, font=\footnotesize},
  model/.style={box, fill=gray!10},
  data/.style={box},
  schema/.style={box, dashed, fill=gray!5},
  artifact/.style={box, fill=gray!20},
  >={Stealth[length=1.4mm]},
  flow/.style={->, >=Stealth, shorten >=1pt, shorten <=1pt, thick},
  inject/.style={->, >=Stealth, dashed, shorten >=1pt, shorten <=1pt},
]
\node[data] (B1) {pattern\_md};
\node[model, right=of B1] (B2) {LLM};
\node[data, right=of B2] (B3) {C\#};
\node[artifact, right=of B3] (B4) {Unity};
\node[data, right=of B4] (B5) {err log};
\node[left=1mm of B1, font=\footnotesize\itshape, anchor=east] {Base};
\draw[flow] (B1) -- (B2); \draw[flow] (B2) -- (B3); \draw[flow] (B3) -- (B4); \draw[flow] (B4) -- (B5);

\node[data, below=6mm of B1] (I1) {pattern\_md};
\node[model, right=of I1] (I2) {LLM$_1$};
\node[data, right=of I2] (I3) {IR};
\node[model, right=of I3] (I4) {LLM$_2$};
\node[data, right=of I4] (I5) {C\#};
\node[artifact, right=of I5] (I6) {Unity};
\node[data, right=of I6] (I7) {err log};
\node[left=1mm of I1, font=\footnotesize\itshape, anchor=east] {IR-cond};
\draw[flow] (I1) -- (I2); \draw[flow] (I2) -- (I3); \draw[flow] (I3) -- (I4); \draw[flow] (I4) -- (I5); \draw[flow] (I5) -- (I6); \draw[flow] (I6) -- (I7);

\node[schema, below=5mm of I2] (S) {schema\\(\texttt{free}/\texttt{min}/\texttt{full})};
\draw[inject] (S) -- (I2);
\end{tikzpicture}%
}
\caption{Baseline and IR-conditioned pipelines. Data (white), LLM (light grey), compile target (dark grey), schema injection (dashed). Both share the same Unity batch-replay harness and error-log aggregation.}
\label{fig:pipeline}
\end{figure}

\subsection{IR v0.2-runtime-evidence}
\label{sec:ir}
The IR mediates between goal pattern description and executable Unity realization, derived from the same Unity project used as the evaluation target to ground it simultaneously in concrete implementation structure and goal pattern semantics.

\paragraph{Schema bootstrapping and freeze.} An initial static draft defined six top-level fields from three hand-authored patterns (Ownership, Delivery, Alignment). Static YAML parsing proved insufficient---critical mechanics emerge from prefab instantiation and runtime script logic---so the schema was extended with \texttt{runtime\_params} and refined through five versioned iterations (\nameref{sec:appendix-ir-iteration}). Frequency analysis across all 26 patterns confirmed all seven top-level fields are Core-tier (100\% coverage); the schema was frozen as v0.2-runtime-evidence prior to any generation runs.

\paragraph{Extraction pipeline.} The IR for each pattern is generated programmatically from static analysis of the Unity scene YAML: PrefabInstance blocks are resolved to canonical asset names via a curated GUID map; MonoBehaviour blocks to C\# class names via a script GUID map; serialized field values are extracted into \texttt{runtime\_params}. Semantic \texttt{links} and \texttt{rules} entries---encoding conditional runtime relations evidenced in script source---are assembled by a pattern-specific configuration function. Three scenes were hand-authored as bootstrapping examples; the remaining 23 were batch-processed.

\paragraph{Schema definition and dual grounding.} The frozen schema defines seven fields: \texttt{scene}, \texttt{objects}, \texttt{scripts}, \texttt{params} (always \texttt{\{\}}), \texttt{runtime\_params}, \texttt{links}, and \texttt{rules}, governed by four hard constraints (per-instance script binding, no aggregate placeholders, required \texttt{evidence\_type} on all rules, conditional relation label convention). The IR carries two complementary grounding layers: a \emph{structural layer} grounded in the concrete project implementation and a \emph{semantic layer} grounded in the goal pattern, providing both the implementation information needed for compilable Unity code and the pattern information needed for correct gameplay instantiation. Full schema and annotated example are in \nameref{sec:appendix-ir}.

\subsection{Models and Configurations}
Two code-specialized open-source models were selected: DeepSeek-Coder-V2-Lite-Instruct~\cite{dai2024deepseekmoe} and Qwen2.5-Coder-7B-Instruct~\cite{hui2024qwen2,qwen2} chosen for code specialization, open weights, and HPC batch inference feasibility via vLLM. In addition, DeepSeek-Coder-V2-Lite-Instruct and Qwen2.5-Coder-7B-Instruct report the highest pass@1 percentage on HumanEval among recent open-source models ~\cite{jiang2026survey,chen2021evaluating}. The evaluation comprises $2\ \text{models} \times 4\ \text{configurations} \times 520\ \text{records} = 4{,}160$ total records (520 = 26 patterns $\times$ 20 seeds), run on NVIDIA A40 GPUs via vLLM~\cite{kwon2025vllm} v0.7.3 under Apptainer \footnote{\url{https://github.com/apptainer/apptainer}}. Decoding is done with the following settings: temperature $= 0.2$, top\_p $= 0.95$, max\_model\_len $= 3072$; \texttt{max\_tokens} $= 2048$ (C\#) and $4096$ (IR).

\begin{table}[t]
\centering
\caption{Prompt differences across configurations. All four share a header (\texttt{[pattern: <PATTERN\_ID>]}, \texttt{[method: <METHOD>]}) and ingest \texttt{<PATTERN\_MD>}. Full verbatim templates: \nameref{sec:appendix-prompt-template}.}
\label{tab:prompts}
\scriptsize
\begin{tabularx}{\linewidth}{@{}l l X@{}}
\toprule
Configuration & Target & Additional prompt content (beyond shared header and \texttt{<PATTERN\_MD>}) \\
\midrule
\texttt{no\_schema}         & C\#               & Direct Unity Editor C\# generation; ``Output only raw C\# code.'' \\
\texttt{with\_schema\_free} & IR JSON $\to$ C\# & IR step: no schema constraints; ``Output ONLY valid JSON.'' \\
\texttt{with\_schema\_min}  & IR JSON $\to$ C\# & IR step: \texttt{free} + skeleton listing the 7 required top-level fields with shape annotations. \\
\texttt{with\_schema\_full} & IR JSON $\to$ C\# & IR step: \texttt{min} + type annotations + 4 hard referential-integrity constraints enumerated inline. \\
\bottomrule
\end{tabularx}
\end{table}

\subsection{Evaluation Protocol}
\label{sec:eval_protocol}
\paragraph{Primary metric: M1 Compile Success.} M1 is binary compile viability from Unity batch replay logs: each generated C\# script is written to a temporary asset file, Unity 2022.2.23f1 is invoked in batchmode\footnote{\url{https://docs.unity3d.com/2019.1/Documentation/Manual/CommandLineArguments.html}}, and the Editor log is scanned for compiler error codes. Records exceeding 120 seconds are terminated and marked \texttt{compile\_timeout}. Compilation is a minimal operational threshold: without it, an artifact cannot exist as an executable candidate.

\paragraph{Failure analysis and reproducibility.} Compiler error codes are extracted from Unity batch logs and analyzed to characterize failure modes; timeout records are reported separately. Pattern data, playable implementations, pipeline code, and reproduction instructions are available in the supplementary repository.\footnote{\url{https://anonymous.4open.science/r/llm-goal-playable-pattern-E312/README.md}}

\section{Results}
\label{sec:results}

\subsection{M1 Compile Success}
No generated artifact achieved successful compilation across either model or any pipeline configuration: pass@$k = 0.0$ for all $k \in \{1, \ldots, 20\}$ across all 26 patterns, two models, and four configurations.

A notable secondary finding is the sharp increase in compilation timeout rate under IR-conditioned configurations. Under no\_schema, timeout rates range from 37.5\% (DeepSeek) to 51.5\% (Qwen); under IR-conditioned configurations they rise monotonically with schema detail, reaching 96--99\% under with\_schema\_full. Compilation timeout is recorded when Unity's BatchRunner watchdog terminates a compilation-and-domain-reload cycle after 120 seconds. The monotonic increase may suggest that IR-conditioned generation produces structurally more complex C\# outputs that systematically exhaust the compilation budget. Error code data for IR-conditioned configurations should therefore be interpreted as partial compilation evidence.

\subsection{Failure Distribution}
\label{sec:failure-dist}

\subsubsection{Observed Compiler Errors}

Table~\ref{tab:error_codes} lists all 41 C\# compiler error codes observed across the evaluation, with their compiler-reported message templates and observed frequencies by configuration.
\begin{table*}
\centering
\caption{All observed C\# compiler error codes, message templates, and log count by configuration. G\,=\,grounding failure; H\,=\,hygiene failure. NS\,=\,no schema. Placeholders \texttt{X}, \texttt{Y}, \texttt{T} stand for observed symbols in the raw compiler output. Representative examples from our logs include pattern-named types (\texttt{EvadeGoal}, \texttt{GuardGoal}, \texttt{RescueGoal.Rescue()}) which appear exclusively in G rows as types hallucinated by the model. Other representative examples include script members (\texttt{currentCount}, \texttt{interferableGoal}), namespace paths (\texttt{UnityEngine.Animations}), and primitive/engine types (\texttt{string}, \texttt{Vector2ArrayField}) which occur in both G and H rows depending on the error class.}
\label{tab:error_codes}
\normalsize
\begin{tabularx}{\linewidth}{@{}l X r r r r r@{}}
\toprule
Code & Compiler message & NS & Free & Min & Full & Total \\
\midrule
CS0029 & Cannot implicitly convert \texttt{X} (e.g., \texttt{string}) to \texttt{Y} (e.g., \texttt{string[]}) &  2 &  1 &  0 &  0 &  3 \\
CS0101 & Namespace already contains a definition for \texttt{X} (e.g., \texttt{PlayerController}) & 25 & 18 & 15 & 22 & 80 \\
CS0103 & The name \texttt{X} (e.g., \texttt{whatToConceal}) does not exist in the current context & 14 & 16 & 14 & 15 & 59 \\
CS0111 & Member \texttt{X} (e.g., \texttt{Update}) already defined with same parameter types & 16 & 14 &  9 & 18 & 57 \\
CS0115 & \texttt{X} (e.g., \texttt{EvadeGoal.IsCompleted()}) is not a suitable method for override & 40 &  0 &  0 &  0 & 40 \\
CS0116 & Namespace cannot directly contain members such as fields or methods &  0 &  0 &  0 &  1 &  1 \\
CS0117 & \texttt{X} (e.g., \texttt{EditorGUILayout}) does not contain a definition for \texttt{Y} (e.g., \texttt{Vector2ArrayField}) &  2 &  2 &  0 &  1 &  5 \\
CS0122 & \texttt{X} (e.g., \texttt{RescueGoal.Rescue()}) is inaccessible due to its protection level & 17 &  0 &  0 &  0 & 17 \\
CS0136 & Local \texttt{X} cannot be declared; name used in enclosing scope &  2 &  0 &  0 &  0 &  2 \\
CS0165 & Use of unassigned local variable \texttt{X} (e.g., \texttt{interferableGoal}) & 13 &  0 &  0 &  0 & 13 \\
CS0234 & Type \texttt{X} (e.g., \texttt{Rigging}) does not exist in namespace \texttt{Y} (e.g., \texttt{UnityEngine.Animations}) & 20 &  0 &  0 &  0 & 20 \\
CS0239 & Cannot override sealed member \texttt{X} (e.g., \texttt{TrackAsset.CreatePlayable()}) &  4 &  0 &  0 &  0 &  4 \\
CS0246 & The type or namespace \texttt{X} (e.g., \texttt{GuardGoal}) could not be found & 33 & 11 & 29 & 31 & 104 \\
CS0263 & Partial declarations of \texttt{X} (e.g., \texttt{GainInformation}) must not specify different base classes & 20 &  0 &  0 &  0 & 20 \\
CS0311 & Type \texttt{X} (e.g., \texttt{ActionDescription}) cannot be used as type parameter \texttt{T} in \texttt{Y} (e.g., \texttt{AddComponent<T>()}) &  1 &  0 &  5 &  1 &  7 \\
CS0315 & Type \texttt{X} (e.g., \texttt{UnityEngine.Color}) cannot be used as type parameter \texttt{T}; no boxing conversion &  0 &  0 &  5 &  0 &  5 \\
CS0509 & Cannot derive from sealed type \texttt{X} (e.g., \texttt{ClipCaps}) &  1 &  0 &  0 &  0 &  1 \\
CS0595 & Invalid real literal &  0 & 10 &  3 &  1 & 14 \\
CS0619 & \texttt{X} (e.g., \texttt{AddComponent(string)}) is obsolete &  0 &  0 &  1 &  0 &  1 \\
CS1001 & Identifier expected &  1 & 24 & 24 & 17 & 66 \\
CS1002 & \texttt{;} expected &  9 &  0 &  3 &  2 & 14 \\
CS1003 & Syntax error, \texttt{X} (e.g., \texttt{,}) expected &  4 & 20 & 22 & 12 & 58 \\
CS1010 & Newline in constant &  6 &  0 &  0 &  0 &  6 \\
CS1012 & Too many characters in character literal &  1 &  0 &  0 &  0 &  1 \\
CS1013 & Invalid number &  0 & 20 & 17 & 11 & 48 \\
CS1022 & Type or namespace definition, or end-of-file expected &  6 &  0 &  3 &  2 & 11 \\
CS1026 & \texttt{)} expected &  2 &  0 &  0 &  0 &  2 \\
CS1029 & \texttt{\#error} directive (e.g., \texttt{BatchRunner\_sanitize\_no\_csharp\_start}) & 19 & 40 & 40 & 40 & 139 \\
CS1040 & Preprocessor directive must be first non-whitespace on line &  0 &  0 &  3 &  1 &  4 \\
CS1041 & Identifier expected; \texttt{X} (e.g., \texttt{in}) is a keyword &  0 &  1 &  1 &  0 &  2 \\
CS1056 & Unexpected character &  2 &  0 &  0 &  0 &  2 \\
CS1061 & \texttt{X} (e.g., \texttt{PlayerController}) does not contain a definition for \texttt{Y} (e.g., \texttt{health}) &  3 &  4 & 10 & 12 & 29 \\
CS1503 & Argument \texttt{N}: cannot convert from \texttt{X} (e.g., \texttt{string}) to \texttt{Y} (e.g., \texttt{UnityEngine.Object}) & 23 &  1 &  2 &  0 & 26 \\
CS1513 & \texttt{\}} expected & 24 &  0 &  3 &  2 & 29 \\
CS1519 & Invalid token in member declaration &  1 &  0 &  0 &  0 &  1 \\
CS1525 & Invalid expression term &  1 &  0 &  0 &  0 &  1 \\
CS1529 & \texttt{using} clause must precede all other elements &  4 & 24 & 26 & 18 & 72 \\
CS1624 & Body of \texttt{X} (e.g., \texttt{MoveGameElement()}) cannot be an iterator block because return type is \texttt{Y} (e.g., \texttt{void}) &  0 &  1 &  0 &  0 &  1 \\
CS2001 & Source file \texttt{X} could not be found &  6 &  9 &  4 &  8 & 27 \\
CS8121 & Expression of type \texttt{X} (e.g., \texttt{Component}) cannot be handled by pattern of type \texttt{Y} (e.g., \texttt{ScriptableObject}) &  0 &  0 &  3 &  0 &  3 \\
CS8803 & Top-level statements must precede namespace declarations &  7 &  0 &  1 &  2 & 10 \\
\midrule
\textit{Total} & & 329 & 216 & 243 & 217 & 1005 \\
\bottomrule
\end{tabularx}
\end{table*}

Inspection of the error messages reveals two distinct failure types. A first group of 13 errors---CS0115, CS0117, CS0122, CS0234, CS0239, CS0246, CS0311, CS0315, CS0509, CS0619, CS1061, CS1624, CS8121---all involve references to types, members, namespaces, or inheritance structures that do not exist in the target Unity project or engine API. These errors indicate that the model generated code assuming the existence of constructs that are absent from the actual codebase. The required knowledge is dual in nature: accurate knowledge of the target project's Unity implementation structure (prefab identifiers, script class names, component bindings) and accurate knowledge of the goal pattern vocabulary (which mechanics and relations the pattern requires). Both layers are encoded in the IR; their absence in the no\_schema condition is precisely what these errors reflect. We term these \emph{grounding failures}.

A second group of 28 errors---CS0029, CS0101, CS0103, CS0111, CS0116, CS0136, CS0165, CS0263, CS0595, CS1001, CS1002, CS1003, CS1010, CS1012, CS1013, CS1022, CS1026, CS1029, CS1040, CS1041, CS1056, CS1503, CS1513, CS1519, CS1525, CS1529, CS2001, CS8803---reflects syntax corruption, duplicate declarations, formatting leakage, and type coercion errors. These errors are independent of the model's knowledge of the project or pattern information: they would occur even if all referenced types and constructs existed. We term these \emph{hygiene failures} — errors independent of grounding that are in principle addressable through constrained decoding or output sanitization — borrowing the notion of hygiene from programming language theory \cite{kohlbecker1986hygienic} and software engineering practice \cite{tilbrook1990washing}.

\subsubsection{Grounding Failures}

Grounding failures (as short as G failures) reflect the model's inability to map goal-pattern semantics onto the actual implementation primitives available in the target Unity project. Under no\_schema, this category accounts for 121 out of 329 total logged error instances (36.8\%). The dominant codes are CS0115 (40 logs), CS0122 (17 logs), CS0234 (20 logs), and CS0246 (33 logs), collectively indicating failures at three grounding layers:

Failures occur at three layers: \textbf{project-level} (CS0122, CS0246; 17 and 33 logs under no\_schema), \textbf{engine API} (CS0117, CS0234, CS0311, CS0315, CS0619, CS1061, CS8121; present across all configurations), and \textbf{architectural} (CS0115, CS0239, CS0509, CS1624; 45 logs under no\_schema, nearly absent under IR-conditioned generation).

The IR-conditioned configurations show a marked reduction in grounding failures relative to no\_schema. Under with\_schema\_free, the grounding-sensitive total falls to 18 (8.3\% of logged errors), driven primarily by the near-elimination of CS0115, CS0234, and CS0122. Under with\_schema\_min and with\_schema\_full, however, grounding failure totals rise to 53 (21.8\%) and 45 (20.7\%) respectively, driven primarily by the persistence of CS0246 (29 and 31 logs respectively) and an increase in CS1061 (10 and 12 logs, compared to 3 under no\_schema). CS0311 and CS0315 also appear under with\_schema\_min (5 logs each) but are absent or near-absent in other configurations. These interpretations must be qualified by the high timeout rates under all IR-conditioned configurations: the majority of records did not reach a complete compilation cycle, and the error code data reflects partial logs only.

\subsubsection{Hygiene Failures}

Hygiene failures (as short as H failures) reflect pipeline hygiene and output formatting issues rather than grounding deficits. Under no\_schema, this category accounts for 208 out of 329 total logged error instances (63.2\%), dominated by duplicate declaration errors (CS0101, CS0111, CS0263), type coercion errors (CS1503, CS1513), and unassigned-local errors (CS0165). The composition shifts markedly under IR-conditioned generation. Under with\_schema\_free, with\_schema\_min, and with\_schema\_full, hygiene failures account for 198 (91.7\%), 190 (78.2\%), and 172 (79.3\%) of logged errors respectively. Codes associated with output formatting and sanitizer rejection---CS1029 (marker comment leakage), CS1001, CS1003, CS1013, and CS1529---become dominant, replacing the duplicate declaration and unassigned-local errors that characterise no\_schema output. This compositional shift indicates progress: surface formatting failures are addressable through post-processing, whereas duplicate declarations and unassigned-local errors reflect deeper generation-side issues. CS1029 saturates at 40 log files across all three IR-conditioned configurations, suggesting systematic failure to strip IR-related markup. CS0263 and CS0165, prominent under no\_schema (20 and 13 logs respectively), are entirely absent under IR-conditioned generation. These interpretations are subject to a caveat: high timeout rates under IR-conditioned configurations mean that compilation errors manifesting late in the build cycle are systematically unobserved, and the true error distribution may differ from partial logs.

\subsubsection{Cross-Model Comparison}

As shown in Table \ref{tab:model_error_breakdown}, under no\_schema, where timeout rates are sufficiently low to support reliable comparison (Qwen: 51.5\%, DeepSeek: 37.5\%), both models produce grounding failures at comparable absolute levels (Qwen: 64, DeepSeek: 57). However, their grounding failure profiles are compositionally distinct: Qwen produces CS0234 (20 vs.\ 0) and CS0239 (4 vs.\ 0), while DeepSeek produces CS0122 (17 vs.\ 0) and CS0246 at a higher rate (20 vs.\ 13). Both models converge on CS0115 (20 each), supporting the interpretation that architectural grounding failure is structural rather than model-specific. Hygiene failures diverge more sharply: DeepSeek shows higher rates of duplicate declaration and structural errors---CS0101 (20 vs.\ 5), CS0263 (20 vs.\ 0), CS0165 (13 vs.\ 0), and CS1513 (20 vs.\ 4)---while Qwen produces CS1029 (19 vs.\ 0) and CS1503 (19 vs.\ 4) at substantially higher rates.

Under with\_schema\_free (timeout 86--92\%), Qwen is dominated by CS1001/CS1003/CS1013/CS1529 while DeepSeek shows higher CS0101/CS0111; CS1029 saturates at 20 log files for both. Under with\_schema\_min and with\_schema\_full (timeout $\geq$96\%), quantitative comparison is unreliable; the most consistent pattern is CS0101/CS0111 persistently higher for DeepSeek and CS0103/CS1013 for Qwen, with CS1029 saturating at 20 per model across all IR-conditioned configurations.

\begin{table}
\centering
\caption{Per-model log\_count by error code and configuration. G\,=\,grounding failure; H\,=\,hygiene failure. NS\,=\,no\_schema, Free\,=\,with\_schema\_free, Min\,=\,with\_schema\_min, Full\,=\,with\_schema\_full.}
\label{tab:model_error_breakdown}
\scriptsize
\begin{tabularx}{\linewidth}{@{}l l rrrr rrrr@{}}
\toprule
 & & \multicolumn{4}{c}{Qwen} & \multicolumn{4}{c}{DeepSeek} \\
\cmidrule(lr){3-6}\cmidrule(lr){7-10}
Class & Code & NS & Free & Min & Full & NS & Free & Min & Full \\
\midrule
\multirow{13}{*}{G}
 & CS0115 & 20 &  0 &  0 &  0 & 20 &  0 &  0 &  0 \\
 & CS0117 &  2 &  2 &  0 &  1 &  0 &  0 &  0 &  0 \\
 & CS0122 &  0 &  0 &  0 &  0 & 17 &  0 &  0 &  0 \\
 & CS0234 & 20 &  0 &  0 &  0 &  0 &  0 &  0 &  0 \\
 & CS0239 &  4 &  0 &  0 &  0 &  0 &  0 &  0 &  0 \\
 & CS0246 & 13 &  3 & 15 & 19 & 20 &  8 & 14 & 12 \\
 & CS0311 &  1 &  0 &  4 &  1 &  0 &  0 &  1 &  0 \\
 & CS0315 &  0 &  0 &  5 &  0 &  0 &  0 &  0 &  0 \\
 & CS0509 &  1 &  0 &  0 &  0 &  0 &  0 &  0 &  0 \\
 & CS0619 &  0 &  0 &  1 &  0 &  0 &  0 &  0 &  0 \\
 & CS1061 &  3 &  0 &  1 & 10 &  0 &  4 &  9 &  2 \\
 & CS1624 &  0 &  1 &  0 &  0 &  0 &  0 &  0 &  0 \\
 & CS8121 &  0 &  0 &  3 &  0 &  0 &  0 &  0 &  0 \\
\midrule
\multirow{28}{*}{H}
 & CS0029 &  2 &  1 &  0 &  0 &  0 &  0 &  0 &  0 \\
 & CS0101 &  5 &  1 &  1 &  4 & 20 & 17 & 14 & 18 \\
 & CS0103 &  7 & 14 & 13 & 15 &  7 &  2 &  1 &  0 \\
 & CS0111 &  6 &  0 &  1 &  2 & 10 & 14 &  8 & 16 \\
 & CS0116 &  0 &  0 &  0 &  0 &  0 &  0 &  0 &  1 \\
 & CS0136 &  0 &  0 &  0 &  0 &  2 &  0 &  0 &  0 \\
 & CS0165 &  0 &  0 &  0 &  0 & 13 &  0 &  0 &  0 \\
 & CS0263 &  0 &  0 &  0 &  0 & 20 &  0 &  0 &  0 \\
 & CS0595 &  0 & 10 &  3 &  0 &  0 &  0 &  0 &  1 \\
 & CS1001 &  1 & 20 & 14 &  9 &  0 &  4 & 10 &  8 \\
 & CS1002 &  2 &  0 &  0 &  0 &  7 &  0 &  3 &  2 \\
 & CS1003 &  2 & 20 & 14 &  9 &  2 &  0 &  8 &  3 \\
 & CS1010 &  1 &  0 &  0 &  0 &  5 &  0 &  0 &  0 \\
 & CS1012 &  1 &  0 &  0 &  0 &  0 &  0 &  0 &  0 \\
 & CS1013 &  0 & 20 & 14 &  9 &  0 &  0 &  3 &  2 \\
 & CS1022 &  1 &  0 &  0 &  0 &  5 &  0 &  3 &  2 \\
 & CS1026 &  0 &  0 &  0 &  0 &  2 &  0 &  0 &  0 \\
 & CS1029 & 19 & 20 & 20 & 20 &  0 & 20 & 20 & 20 \\
 & CS1040 &  0 &  0 &  0 &  0 &  0 &  0 &  3 &  1 \\
 & CS1041 &  0 &  1 &  0 &  0 &  0 &  0 &  1 &  0 \\
 & CS1056 &  2 &  0 &  0 &  0 &  0 &  0 &  0 &  0 \\
 & CS1503 & 19 &  0 &  2 &  0 &  4 &  1 &  0 &  0 \\
 & CS1513 &  4 &  0 &  0 &  0 & 20 &  0 &  3 &  2 \\
 & CS1519 &  1 &  0 &  0 &  0 &  0 &  0 &  0 &  0 \\
 & CS1525 &  1 &  0 &  0 &  0 &  0 &  0 &  0 &  0 \\
 & CS1529 &  4 & 20 & 14 &  9 &  0 &  4 & 12 &  9 \\
 & CS2001 &  2 &  4 &  3 &  4 &  4 &  5 &  1 &  4 \\
 & CS8803 &  2 &  0 &  0 &  0 &  5 &  0 &  1 &  2 \\
\bottomrule
\end{tabularx}
\end{table}

\subsubsection{Pattern-Level Failure Distribution}
\label{sec:pattern-failures}

Tables~\ref{tab:pattern_all} aggregate both models (40 logs per pattern per configuration (20 seeds $\times$ 2 models)). Aggregation is appropriate at the configuration level given the cross-model consistency in grounding failure categories. However, pattern-level grounding failure counts in some cases are model-specific rather than shared behaviour; per-model breakdown is provided in~\nameref{sec:appendix-pattern-bymodel}.

\begin{table*}
\centering
\caption{Pattern-level errors by configuration (both models combined, 40 logs per pattern). G\,=\,grounding failure; H\,=\,hygiene failure. Timeout rates: NS 37--51\%, Free 86--92\%, Min 96\%, Full 97--99\%.}
\label{tab:pattern_all}
\scriptsize
\begin{tabularx}{\textwidth}{@{}l rrr X rrr X rrr X rrr X@{}}
\toprule
& \multicolumn{4}{c}{no\_schema} & \multicolumn{4}{c}{with\_schema\_free} & \multicolumn{4}{c}{with\_schema\_min} & \multicolumn{4}{c}{with\_schema\_full} \\
\cmidrule(lr){2-5}\cmidrule(lr){6-9}\cmidrule(lr){10-13}\cmidrule(lr){14-17}
Pattern & Tot & G & H & Top G & Tot & G & H & Top G & Tot & G & H & Top G & Tot & G & H & Top G \\
\midrule
1\_Ownership       & 18 &  0 & 18 & ---                        &  31 &  9 & 22 & CS0246(6), CS1061(3)       &  71 & 14 & 57 & CS0246(11), CS1061(3)      &  49 & 16 & 33 & CS0246(15), CS0117(1) \\
2\_Collection      & 23 &  8 & 15 & CS0115(2), CS0234(2)       &  47 &  0 & 47 & ---                        &  50 &  0 & 50 & ---                        &  47 &  8 & 39 & CS0246(6), CS1061(2) \\
3\_Eliminate       & 37 &  0 & 37 & ---                        &  46 &  2 & 44 & CS0117(2)                  &  43 &  7 & 36 & CS0246(4), CS8121(2)       &  43 &  0 & 43 & --- \\
4\_Capture         & 41 & 27 & 14 & CS0234(16), CS0115(11)     &  47 &  0 & 47 & ---                        &  38 &  0 & 38 & ---                        &  35 &  0 & 35 & --- \\
5\_Overcome        &  5 &  1 &  4 & CS0234(1)                  &  25 &  0 & 25 & ---                        &  44 & 15 & 29 & CS0246(7), CS1061(5)       &  54 & 10 & 44 & CS0246(5), CS1061(5) \\
6\_Evade           & 50 & 32 & 18 & CS0115(16), CS0246(15)     &  46 &  0 & 46 & ---                        &  38 &  4 & 34 & CS0246(2), CS1061(2)       &  37 &  0 & 37 & --- \\
7\_Stealth         & 52 & 50 &  2 & CS0115(24), CS0246(19)     &  41 &  0 & 41 & ---                        &  36 &  0 & 36 & ---                        &  44 &  6 & 38 & CS0246(3), CS1061(3) \\
8\_Herd\_Attract   & 24 &  6 & 18 & CS0115(3), CS0234(2)       &  43 &  1 & 42 & CS0246(1)                  &  36 &  1 & 35 & CS0246(1)                  &  36 &  0 & 36 & --- \\
9\_Conceal         & 53 &  7 & 46 & CS0115(3), CS0239(2)       &  47 &  0 & 47 & ---                        &  40 &  0 & 40 & ---                        &  44 &  0 & 44 & --- \\
10\_Rescue         & 31 & 22 &  9 & CS0122(17), CS0246(5)      &  45 &  0 & 45 & ---                        &  49 &  0 & 49 & ---                        &  46 &  0 & 46 & --- \\
11\_Delivery       & 54 & 27 & 27 & CS0234(16), CS0115(11)     &  63 &  0 & 63 & ---                        &  44 &  4 & 40 & CS0246(4)                  &  41 &  0 & 41 & --- \\
12\_Guard          & 17 & 16 &  1 & CS0234(8), CS0115(7)       &  41 &  0 & 41 & ---                        &  45 &  0 & 45 & ---                        &  36 &  0 & 36 & --- \\
13\_Race           &  9 &  6 &  3 & CS0234(4), CS0115(1)       &  26 &  2 & 24 & CS0246(2)                  &  36 &  4 & 32 & CS0246(3), CS1061(1)       &  27 &  4 & 23 & CS0246(4) \\
14\_Alignment      & 21 &  5 & 16 & CS0234(3), CS0115(2)       &  37 &  0 & 37 & ---                        &  47 &  0 & 47 & ---                        &  55 & 10 & 45 & CS0246(5), CS1061(4) \\
15\_Configuration  &  5 &  1 &  4 & CS0246(1)                  &  45 &  0 & 45 & ---                        &  45 &  9 & 36 & CS0246(5), CS0315(4)       &  56 &  0 & 56 & --- \\
16\_Traverse       & 33 & 21 & 12 & CS0234(19), CS0115(2)      &  32 &  1 & 31 & CS1061(1)                  &  36 &  0 & 36 & ---                        &  41 &  0 & 41 & --- \\
17\_Survive        &  9 &  7 &  2 & CS0115(4), CS0234(2)       &  41 &  1 & 40 & CS1624(1)                  &  43 & 11 & 32 & CS0246(8), CS0311(2)       &  38 &  0 & 38 & --- \\
18\_Connection     & 20 &  6 & 14 & CS0115(2), CS0234(2)       &  37 &  0 & 37 & ---                        &  48 &  1 & 47 & CS0246(1)                  &  39 &  0 & 39 & --- \\
19\_Exploration    &  4 &  0 &  4 & ---                        &  61 &  0 & 61 & ---                        &  38 &  3 & 35 & CS0246(1), CS0311(1)       &  40 &  3 & 37 & CS0246(3) \\
20\_Reconnaissance & 11 &  6 &  5 & CS0246(6)                  &  44 &  2 & 42 & CS0246(2)                  &  37 &  1 & 36 & CS1061(1)                  &  35 &  3 & 32 & CS0246(2), CS1061(1) \\
21\_Contact        & 30 & 28 &  2 & CS0246(11), CS0234(9)      &  53 &  0 & 53 & ---                        &  37 &  0 & 37 & ---                        &  35 &  5 & 30 & CS0246(5) \\
22\_Enclosure      &  7 &  5 &  2 & CS0246(2), CS0115(1)       &  53 &  0 & 53 & ---                        &  46 &  2 & 44 & CS0246(2)                  &  46 &  1 & 45 & CS0246(1) \\
23\_GainCompetence &  5 &  2 &  3 & CS0246(1), CS0311(1)       &  44 &  0 & 44 & ---                        &  38 &  0 & 38 & ---                        &  42 &  1 & 41 & CS0246(1) \\
24\_GainInformation& 55 & 10 & 45 & CS0239(4), CS0115(3)       &  62 &  0 & 62 & ---                        &  43 &  4 & 39 & CS0246(4)                  &  41 &  0 & 41 & --- \\
25\_LastManStanding& 15 &  0 & 15 & ---                        &  44 &  0 & 44 & ---                        &  37 &  2 & 35 & CS0246(2)                  &  35 &  0 & 35 & --- \\
26\_KingoftheHill  & 37 &  4 & 33 & CS0246(2), CS0115(1)       &  35 &  0 & 35 & ---                        &  38 &  0 & 38 & ---                        &  33 &  0 & 33 & --- \\
\bottomrule
\end{tabularx}
\end{table*}

Across all configurations, G failures are distributed unevenly across patterns. Under no\_schema, 22 of 26 patterns exhibit at least one G failure; 4 patterns (1\_Ownership, 3\_Eliminate, 19\_Exploration, 25\_LastManStanding) show zero G failures with all logged failures attributable to H causes. Under with\_schema\_free, G failures are reduced to 7 patterns, with CS1061, CS0117, CS0246 and CS1624 as the only remaining G codes. Under with\_schema\_min and with\_schema\_full, G failures reappear across more patterns, with CS0246 dominant in nearly all affected cases; however, these distributions are based on partial logs due to high timeout rates and should be interpreted with caution.

A consistent cross-configuration observation is that 1\_Ownership exhibits persistent G failures across all three setups (G log counts: 9, 14, 16 under with\_schema\_free, min, and full respectively), despite showing zero G failures under no\_schema, in contrast to structurally similar patterns such as 2\_Collection and 3\_Eliminate which show G failures only under no\_schema or not at all. The interpretation of this cross-configuration pattern is discussed in Section~\nameref{sec:discussion}.

\section{Discussion}
\label{sec:discussion}

\paragraph{IR as representational scaffold.}
IR does not replace generative exploration; it constrains and channels it. The data reveal an asymmetric grounding effect: IR conditioning nearly eliminates architectural grounding failures (CS0115: 40 logs under no\_schema, 0 under all IR-conditioned configurations), confirming that the IR's structural layer successfully transfers MonoBehaviour and inheritance conventions. However, CS0246 (hallucinated project-specific types) persists across all configurations (no\_schema: 33, free: 11, min: 29, full: 31), suggesting that project-level grounding is not fully resolved by schema conditioning alone and may require richer or more targeted knowledge injection. The \texttt{full} configuration's explicit hard constraints did not eliminate grounding failures, and the monotonic increase in compilation timeout with schema detail (37--51\% under no\_schema to 97--99\% under with\_schema\_full) suggests a further problem: IR-conditioned generation produces structurally more complex C\# outputs that systematically exhaust the Unity compilation budget. The IR is simultaneously necessary for grounding and costly for compilation tractability. This tension---too sparse for reliable grounding without it, too complex for reliable compilation with it---defines the current boundary of the approach.

\paragraph{Pattern-level grounding demands.}
Inspection of the per-log top-G codes isolates a consistent anomaly: Ownership's residual G failures are dominated by CS0246 (6, 11, 15 occurrences under Free/Min/Full) and CS1061 (3, 3, 0), indicating the model is referring to types and members that do not exist in the target project.

The semantic source is visible in the reference IR itself: Ownership is the only pattern in the 26-pattern set whose win condition is mediated by a \textbf{per-instance state-change relation}---each \texttt{OwnershipObject} prefab instance must have its colour changed on trigger \textbf{and} increment a separate \texttt{GoalManager.currentCount}---which requires the IR-consuming LLM to name both the per-instance script (\texttt{ChangeColor}) and a specific runtime member (\texttt{currentCount}) not expressible through \texttt{GameManager}/\texttt{SpawnManager}/\texttt{GoalManager} alone.

Under no-schema, the model avoids this hazard by falling back to surface-level boilerplate (H failures dominate, total 18). Under IR conditioning, the schema obliges the model to commit to concrete per-instance script and member identifiers, and it hallucinates plausible-but-absent ones.

Collection and Eliminate, by contrast, route their win condition through a single aggregate counter on \texttt{GoalManager}, for which the schema already supplies canonical identifiers.

This anomaly generalises an observation visible throughout the pattern-level tables: the patterns that persist in showing G failures under IR conditioning (Ownership, Overcome, Survive, Alignment, Configuration) are precisely those whose rules reference script-defined per-instance state, rather than aggregate \texttt{GoalManager} counters.

The IR's structural layer transfers project-level identifier vocabulary but not the \textbf{discipline} of choosing between aggregate and per-instance mediation---the latter is pattern-specific semantic knowledge that the current schema does not externalise.

This suggests that the productive next step is not a broader schema but a pattern-level enrichment that names, for each pattern, the mediation path through which its win condition is realised.

\paragraph{Human-machine division of labor.}
Human-machine co-creativity research distinguishes between systems where humans and machines occupy complementary generative roles, each contributing what the other cannot efficiently provide~\cite{liapis2016mixed,deterding2017mixed}. Our pipeline instantiates this division explicitly: human authors contribute domain knowledge, conceptual structure, and representational schema grounded in expert game design practice; the model contributes generative breadth across the space of possible realizations. Neither role is substitutable by the other---the IR cannot be automatically derived without human design knowledge, and the scale of instantiation cannot be achieved through manual authoring alone.

The current results reveal an asymmetry in this division: the human-authored schema successfully encodes \emph{what} exists in the project, but does not yet fully specify \emph{how} Unity's architectural conventions govern the use of those elements. Pattern-level analysis further reveals that grounding difficulty is not uniformly distributed: under no\_schema, grounding failure counts range from 0 to 50 across patterns, suggesting that some goal patterns impose systematically higher grounding demands than others. This suggests that the productive boundary between human and machine contribution may need to be located at the pattern level rather than uniformly across the task space---some patterns may require richer or more targeted knowledge transfer than others before reliable machine-side realization becomes possible. Co-creative system design, in this framing, is not only a question of task allocation but of \emph{knowledge boundary negotiation} grounded in the specific demands of individual design patterns.

The immediate next step is not to improve the IR schema but to relocate the human/machine boundary: either through constrained decoding that enforces syntactic hygiene on the machine side, or through scene-level generation that sidesteps the full compilation problem altogether. In either case, the failure taxonomy established here provides the diagnostic foundation for that relocation---identifying not just that creative realization fails, but precisely where and why.

\paragraph{Future directions.}
The present work establishes compile-grounded viability as a necessary precondition for deeper creative evaluation; future work will extend toward structural adherence, gameplay meaning preservation, and near-pattern confusion analysis (e.g., distinguishing Stealth from Rescue~\cite{lyu2023goal}), with human intervention requirements as a further dimension of computational creativity assessment. The failure taxonomy identifies two orthogonal intervention targets addressable independently: grounding failures via a GNN embedding over the gameplay design pattern graph~\cite{peng2023knowledge}, injectable via PEFT~\cite{liu2023pre,gururangan2020don} or RAG~\cite{lewis2020retrieval}; hygiene failures via constrained decoding~\cite{ma-hu-2025-logically,zarriess2021decoding,geng2023grammar} or rule-based sanitization. Separating the two inference steps---NL$\to$IR for semantic interpretation and IR$\to$C\# for syntactic realization---may better match model capability to task demand. The monotonic timeout increase further suggests flattening the prefab layer into explicit enumerable structures, reframing goal-pattern realization as a \emph{single scene generation} problem and shifting the generative challenge from syntactic code correctness to structured compositional assembly.

\paragraph{Limitations and scope.}
Pattern-level model asymmetry (Section~\nameref{sec:pattern-failures}) is reported as an observation; its interpretation requires per-pattern semantic analysis beyond the current scope. Broader claims regarding comparative model performance, configuration optimality, or semantic fidelity remain future work. The evaluation is scoped to a single engine (Unity) and a single pattern category (goal patterns); generalization to other engines or pattern types is not claimed.

\section{Conclusion}

We frame goal-pattern instantiation as a creative realization problem: converting structured game design knowledge into executable digital artifacts under real engine constraints. Using 26 Unity reference instantiations, we establish an execution-grounded evaluation pipeline and analyze where creative realization succeeds or fails across grounding and hygiene failure modes. IR-conditioned generation provides a principled representational interface for knowledge injection, encoding both project-level structural conventions and goal-pattern semantic knowledge derived from human-authored GPC implementations. Uniform compile failure across all configurations reveals that project-level and engine-level grounding remain primary bottlenecks for knowledge-conditioned LLM generation in Unity, establishing the analytical foundation for deeper structural and semantic evaluation of knowledge-conditioned generative game systems.

\section{Acknowledgement}
The batch compilation pipeline adapts the write-to-asset approach introduced in AICommand by Keijiro Takahashi~\footnote{\url{https://github.com/keijiro/AICommand}}.

We thank Staffan Björk and Jussi Holopainen for their input on goal playable concepts and related background.

This work was supported by the Wallenberg AI, Autonomous Systems and Software Program – Humanity and Society (WASP-HS).

The computations and data handling were enabled by resources provided by the National Academic Infrastructure for Supercomputing in Sweden (NAISS), partially funded by the Swedish Research Council through grant agreement no. 2022-06725.

\clearpage

\bibliographystyle{iccc}
\bibliography{iccc}

\clearpage

\appendix

\section{Appendix: IR iteration record}

See Table~\ref{tab:ir_iterations}

\label{sec:appendix-ir-iteration}
\begin{table*}[t]
\centering
\caption{IR schema iteration history from initial draft to frozen v0.2-runtime-evidence.}
\label{tab:ir_iterations}
\small
\begin{tabularx}{\linewidth}{@{}c p{0.8cm} X X X@{}}
\toprule
\# & Iteration & What Changed & Why & Impact \\
\midrule
1 & \textbf{v0 static draft} &
Defined six top-level fields: \texttt{scene}, \texttt{objects}, \texttt{scripts}, \texttt{params}, \texttt{links}, \texttt{rules}. &
Minimal structured representation between pattern description and Unity code generation. &
Established baseline schema consumed by all pipeline stages. \\
\addlinespace
2 & \textbf{MVP narrowing} &
Deferred \texttt{params} extraction; pipeline operates on \texttt{objects}, \texttt{scripts}, \texttt{links}, \texttt{rules} only. \texttt{params} emitted as \texttt{\{\}}. &
\texttt{params} requires GUID resolution (scene $\to$ prefab $\to$ script $\to$ \texttt{.cs}), blocking the initial pipeline. &
Unblocked end-to-end generation without serialized-field extraction. \\
\addlinespace
3 & \textbf{Runtime extension} &
Added \texttt{PrefabInstance}/\texttt{PrefabAsset} object types, script-defined \texttt{rules}, and \texttt{runtime\_params} field. &
Static \texttt{.unity} parsing misses prefab-driven gameplay; core behaviour emerges from prefab instantiation and runtime script logic. &
IR captures the actual gameplay loop; generation can reason about spawned entities and runtime configuration. \\
\addlinespace
4 & \textbf{Per-instance constraint} &
\texttt{scripts[].object\_id} must reference a real \texttt{objects[].id}. No implicit aggregate placeholders (e.g.\ \texttt{circle\_all}). &
Aggregate placeholders create ambiguous references unresolvable during code generation or evaluation. &
Enforces 1:1 script-to-object binding; enables automatic referential integrity validation. \\
\addlinespace
5 & \textbf{Evidence-aware semantics} &
Conditional relation labels (e.g.\ \texttt{can\_trigger\_game\_win\_if\_aligned}). Required \texttt{evidence\_type} on every \texttt{rules[]} entry. Optional \texttt{confidence} field. &
Unconditional labels over-assert determinism for conditional code paths. Evidence attribution improves trust calibration in generated output. &
Generation produces more accurate causal claims; evaluation can filter or weight rules by evidence type and confidence. \\
\bottomrule
\end{tabularx}
\end{table*}

\section{Appendix: Full 26-Pattern Set with IR Statistics for Freezing Schema}
\label{sec:appendix-26-patterns}

See Table ~\ref{tab:ir_frequency} and Table ~\ref{tab:appendix-26-patterns}

\begin{table}[t]
\centering
\caption{IR schema frequency tiers across all 26 patterns (v2).
Thresholds: Core $\geq 80\%$, Common $\geq 40\%$, Optional $< 40\%$.}
\label{tab:ir_frequency}
\small
\begin{tabularx}{\linewidth}{@{}l X r@{}}
\toprule
Tier & Item & Coverage \\
\midrule
\multicolumn{3}{@{}l}{\textit{Top-level fields}} \\
Core     & \texttt{scene}, \texttt{objects}, \texttt{scripts}, \texttt{params}, \texttt{runtime\_params}, \texttt{links}, \texttt{rules} & 26/26 \\
\addlinespace
\multicolumn{3}{@{}l}{\textit{Object types}} \\
Core     & \texttt{GameObject}, \texttt{PrefabInstance} & 26/26 \\
Optional & \texttt{PrefabAsset}                         & 4/26  \\
\addlinespace
\multicolumn{3}{@{}l}{\textit{Script classes}} \\
Core     & \texttt{GameManager}                         & 26/26 \\
Common   & \texttt{SpawnManager}                        & 19/26 \\
Common   & \texttt{GoalManager}                         & 18/26 \\
\addlinespace
\multicolumn{3}{@{}l}{\textit{Link relations}} \\
Core     & \texttt{has\_prefab\_instance}, \texttt{has\_component} & 26/26 \\
Optional & All pattern-specific relations               & $\leq 7/26$ \\
\addlinespace
\multicolumn{3}{@{}l}{\textit{Rule types}} \\
Core     & \texttt{win\_condition}                      & 26/26 \\
Common   & \texttt{trigger\_count}                      & 12/26 \\
\addlinespace
\multicolumn{3}{@{}l}{\textit{Runtime param keys}} \\
Common   & \texttt{spawnStart}, \texttt{spawnCount}, \texttt{spawnRepeat}, \texttt{spawnRangeX}, \texttt{spawnRangeY} & 18--19/26 \\
Common   & \texttt{goalCount}, \texttt{setGoal}, \texttt{currentCount} & 17--18/26 \\
Optional & All pattern-specific keys                    & $\leq 4/26$ \\
\addlinespace
\multicolumn{3}{@{}l}{\textit{Rule evidence types}} \\
Core     & \texttt{direct\_code}                        & 26/26 \\
\bottomrule
\end{tabularx}
\end{table}

\begin{table}[t]
\centering
\caption{Full set of 26 goal patterns with IR structural statistics (v0.2-runtime-evidence). All 26 scenes pass referential integrity and evidence-type validation.}
\label{tab:appendix-26-patterns}
\small
\begin{tabularx}{\linewidth}{@{}X r r r r@{}}
\toprule
Pattern & Objects & Scripts & Links & Rules \\
\midrule
1 Ownership               &  9 &  4 & 12 & 3 \\
2 Collection              &  7 &  4 & 13 & 3 \\
3 Eliminate               &  7 &  5 & 13 & 3 \\
4 Capture                 &  7 &  6 & 14 & 3 \\
5 Overcome                &  8 &  5 & 14 & 4 \\
6 Evade                   & 10 &  7 & 16 & 3 \\
7 Stealth                 & 16 & 10 & 22 & 3 \\
8 Herd/Attract            & 18 & 16 & 35 & 3 \\
9 Conceal                 & 19 & 13 & 28 & 3 \\
10 Rescue                 & 18 & 13 & 29 & 4 \\
11 Delivery               & 13 & 16 & 35 & 3 \\
12 Guard                  & 19 & 29 & 60 & 3 \\
13 Race                   & 19 &  7 & 18 & 3 \\
14 Alignment              & 13 & 11 & 29 & 4 \\
15 Configuration          & 28 & 11 & 23 & 3 \\
16 Traverse               & 19 &  7 & 18 & 3 \\
17 Survive                & 10 &  6 & 11 & 4 \\
18 Connection             &  5 &  3 &  7 & 2 \\
19 Exploration            &  9 &  3 & 11 & 2 \\
20 Reconnaissance         & 10 &  6 & 15 & 3 \\
21 Contact                &  7 &  4 & 10 & 3 \\
22 Enclosure              & 10 &  8 & 17 & 3 \\
23 Gain Competence        &  6 &  3 &  8 & 2 \\
24 Gain Information       &  9 &  6 & 14 & 3 \\
25 Last Man Standing      & 20 &  9 & 23 & 3 \\
26 King of the Hill       & 16 &  9 & 23 & 3 \\
\midrule
\textit{Total}            & 321 & 221 & 533 & 78 \\
\bottomrule
\end{tabularx}
\end{table}

\section*{Appendix: IR v0.2-runtime-evidence Summary}
\label{sec:appendix-ir}

\textbf{Top-level fields (all required).}
\begin{lstlisting}
scene: string
objects: [{ id, name, type }, ...]
scripts: [{ id, object_id, class_name }, ...]
params: {}
runtime_params: { "<script_id>": { ... }, ... }
links: [{ source, target, relation, evidence_type? }, ...]
rules: [{ id, type, description, pattern, evidence_type, confidence? }, ...]
\end{lstlisting}

\textbf{Hard constraints.}
\begin{lstlisting}
1) scripts[].object_id must reference objects[].id
2) scripts are per-instance (no sharing across objects)
3) no implicit aggregate placeholders
4) rules[].evidence_type is required
   in { direct_code, scene_override, inferred }
\end{lstlisting}

\textbf{Annotated example: 1\_Ownership.}
The following is the reference IR extracted from the Unity project for the Ownership goal pattern. Structural fields (\texttt{objects}, \texttt{scripts}, \texttt{runtime\_params}) are derived from static scene YAML analysis; semantic fields (\texttt{links} relations, \texttt{rules}) are hand-authored to encode runtime behaviour evidenced in script source code.

\begin{lstlisting}[basicstyle=\ttfamily\footnotesize]
{
  "scene": "1_Ownership",
  "objects": [
    { "id": "118191953",
      "name": "Canvas",         "type": "GameObject" },
    { "id": "519420028",
      "name": "Main Camera",    "type": "GameObject" },
    { "id": "1570331856",
      "name": "Text (Legacy)",  "type": "GameObject" },
    { "id": "1012039051484866332",
      "name": "Goal Manager",   "type": "PrefabInstance" },
    { "id": "1112099645",
      "name": "Player",         "type": "PrefabInstance" },
    { "id": "775309098",
      "name": "Boundary",       "type": "PrefabInstance" },
    { "id": "8743824104122491932",
      "name": "Game Manager",   "type": "PrefabInstance" },
    { "id": "9011082862537914474",
      "name": "Spawn Manager",  "type": "PrefabInstance" },
    { "id": "prefab_057536c2a19bd9e4b8cdb1cb044a64f1",
      "name": "OwnershipObject","type": "PrefabAsset" }
  ],
  "scripts": [
    { "id": "script_da1b...",
      "object_id": "9011082862537914474",
      "class_name": "SpawnManager" },
    { "id": "script_74fe...",
      "object_id": "1012039051484866332",
      "class_name": "GoalManager" },
    { "id": "script_bf0c...",
      "object_id": "prefab_057536c2a19bd9e4b8cdb1cb044a64f1",
      "class_name": "ChangeColor" },
    { "id": "script_game_manager",
      "object_id": "8743824104122491932",
      "class_name": "GameManager" }
  ],
  "params": {},
  "runtime_params": {
    "script_da1b...": {
      "spawnStart": true,
      "spawnCount": 8,
      "spawnRepeat": false,
      "spawnPrefabGuid": "057536c2..."
    },
    "script_74fe...": {
      "goalCount": 8,
      "setGoal": true
    }
  },
  "links": [
    { "source": "scene",
      "target": "9011082862537914474",
      "relation": "has_prefab_instance" },
    { "source": "9011082862537914474",
      "target": "script_da1b...",
      "relation": "has_component" },
    { "source": "script_da1b...",
      "target": "prefab_057536c2...",
      "relation": "spawns_prefab",
      "evidence_type": "direct_code" },
    { "source": "script_bf0c...",
      "target": "script_74fe...",
      "relation": "increments_current_count_on_trigger",
      "evidence_type": "direct_code" },
    { "source": "script_74fe...",
      "target": "script_game_manager",
      "relation": "can_trigger_game_win_if_count_met",
      "evidence_type": "direct_code" }
  ],
  "rules": [
    { "id": "rule_spawn_on_start",
      "type": "spawn",
      "description": "SpawnManager spawns OwnershipObject
        prefab on Start when spawnStart is true.",
      "pattern": "Ownership",
      "evidence_type": "direct_code",
      "confidence": 1.0 },
    { "id": "rule_collect_changes_color_and_counts",
      "type": "trigger_count",
      "description": "When Player enters OwnershipObject
        trigger, object color is set to player color
        and GoalManager.currentCount increases by 1.",
      "pattern": "Ownership",
      "evidence_type": "direct_code",
      "confidence": 1.0 },
    { "id": "rule_win_when_count_reaches_goal",
      "type": "win_condition",
      "description": "If GoalManager.setGoal is true
        and currentCount equals goalCount,
        GameManager.GameWin() is called.",
      "pattern": "Ownership",
      "evidence_type": "direct_code",
      "confidence": 1.0 }
  ]
}
\end{lstlisting}

\section*{Appendix: Prompt Templates (Verbatim)}
\label{sec:appendix-prompt-template}
\subsection*{No-schema prompt}
\begin{lstlisting}
[pattern: <PATTERN_ID>]
[method: no_schema]

Generate a Unity Editor script that implements
the playable concept described below.
Output only raw C# code.

<PATTERN_MD>
\end{lstlisting}

\subsection*{With-schema prompt (IR to C\#)}
\begin{lstlisting}
[pattern: <PATTERN_ID>]
[method: <METHOD>]

Generate a Unity Editor script that instantiates
a scene matching the following engine-specific
Intermediate Representation (IR). Thereafter,
you may refer to it as IR.
Output only raw C# code.

<IR_JSON>
\end{lstlisting}

\subsection*{IR maker prompt (free)}
\begin{lstlisting}
[pattern: <PATTERN_ID>]
[method: with_schema_free]

Generate an engine-specific Intermediate Representation (IR) JSON for the
playable concept described below. Thereafter, you may refer to it as IR.
Output ONLY valid JSON. No extra text.

<PATTERN_MD>
\end{lstlisting}

\subsection*{IR maker prompt (min skeleton)}
\begin{lstlisting}
[pattern: <PATTERN_ID>]
[method: with_schema_min]

Generate an engine-specific Intermediate Representation (IR) JSON for the
playable concept described below. Thereafter, you may refer to it as IR.
Output ONLY valid JSON. No extra text.

Required top-level fields:
  "scene"          -- string
  "objects"        -- [ { "id", "name", "type" }, ... ]
  "scripts"        -- [ { "id", "object_id", "class_name" }, ... ]
  "params"         -- {}
  "runtime_params" -- { "<script_id>": { ... }, ... }
  "links"          -- [ { "source", "target", "relation" }, ... ]
  "rules"          -- [ { "id", "type", "description", "pattern", "evidence_type" }, ... ]

<PATTERN_MD>
\end{lstlisting}

\subsection*{IR maker prompt (full schema)}
\begin{lstlisting}
[pattern: <PATTERN_ID>]
[method: with_schema_full]

Generate an engine-specific Intermediate Representation (IR) JSON for the
playable concept described below. Thereafter, you may refer to it as IR.
Output ONLY valid JSON. No extra text.

Follow the IR v0.2-runtime-evidence schema precisely.

Top-level fields (all required):
  "scene"          -- string, scene identifier
  "objects"        -- array of { "id", "name", "type" }
                     type in { "GameObject", "PrefabInstance", "PrefabAsset" }
  "scripts"        -- array of { "id", "object_id", "class_name" }
                     one entry per component instance on one object
  "params"         -- always {}
  "runtime_params" -- object keyed by scripts[].id; values are flat { field: value } maps
  "links"          -- array of { "source", "target", "relation", "evidence_type"? }
  "rules"          -- array of { "id", "type", "description", "pattern",
                     "evidence_type", "confidence"? }

Hard constraints:
  1. Every scripts[].object_id MUST reference a real objects[].id (no dangling refs).
  2. Scripts are per-instance; no shared script entries across objects.
  3. Every entity must be listed explicitly in objects (no aggregate placeholders).
  4. Every rules[] entry MUST include evidence_type
     in { "direct_code", "scene_override", "inferred" }.

<PATTERN_MD>
\end{lstlisting}

\section{Appendix: Pattern-Level Error Distribution by Model}
\label{sec:appendix-pattern-bymodel}

Tables~\ref{tab:pattern_ns_ds}--\ref{tab:pattern_full_qwen} provide per-model breakdowns of G and H failure counts
for all four configurations.

\begin{table}[h]
\centering
\caption{Pattern-level errors: no\_schema, DeepSeek-Coder-V2-Lite
(timeout 37--51\%; 20 logs per pattern).
G = grounding failure; H = hygiene failure}
\label{tab:pattern_ns_ds}
\small
\begin{tabularx}{\linewidth}{@{}l r r r X@{}}
\toprule
Pattern & Total & G & H & Top 2 G codes \\
\midrule
1\_Ownership                   &  18 &   0 &  18 & --- \\
2\_Collection                  &   0 &   0 &   0 & --- \\
3\_Eliminate                   &  36 &   0 &  36 & --- \\
4\_Capture                     &  12 &   0 &   12 & --- \\
5\_Overcome                    &   4 &   0 &   4 & --- \\
6\_Evade                       &  30 &  30 &   0 & CS0115(15), CS0246(15) \\
7\_Stealth                     &  38 &  38 &   0 & CS0115(19), CS0246(19) \\
8\_Herd\_Attract               &   2 &   2 &   0 & CS0115(1), CS0246(1) \\
9\_Conceal                     &  40 &   0 &  40 & --- \\
10\_Rescue                     &  17 &  17 &   0 & CS0122(17) \\
11\_Delivery                   &  27 &   0 &  27 & --- \\
12\_Guard                      &   1 &   0 &   1 & --- \\
13\_Race                       &   0 &   0 &   0 & --- \\
14\_Alignment                  &   0 &   0 &   0 & --- \\
15\_Configuration              &   2 &   0 &   2 & --- \\
16\_Traverse                   &  12 &   0 &   12 & --- \\
17\_Survive                    &   1 &   0 &   1 & --- \\
18\_Connection                 &   2 &   0 &   2 & --- \\
19\_Exploration                &   1 &   0 &   1 & --- \\
20\_Reconnaissance             &   1 &   0 &   1 & --- \\
21\_Contact                    &   9 &   9 &   0 & CS0246(9) \\
22\_Enclosure                  &   2 &   2 &   0 & CS0246(2) \\
23\_GainCompetence             &   2 &   0 &   2 & --- \\
24\_GainInformation            &  36 &   0 &  36 & --- \\
25\_LastManStanding            &   2 &   0 &   2 & --- \\
26\_KingoftheHill              &  17 &   0 &  17 & --- \\
\bottomrule
\end{tabularx}
\end{table}

\begin{table}[h]
\centering
\caption{Pattern-level errors: no\_schema, Qwen2.5-Coder-7B
(timeout 37--51\%; 20 logs per pattern).
G = grounding failure; H = hygiene failure}
\label{tab:pattern_ns_qwen}
\small
\begin{tabularx}{\linewidth}{@{}l r r r X@{}}
\toprule
Pattern & Total & G & H & Top 2 G codes \\
\midrule
1\_Ownership                   &   0 &   0 &   0 & --- \\
2\_Collection                  &  23 &   8 &  15 & CS0115(2), CS0234(2) \\
3\_Eliminate                   &   1 &   0 &   1 & --- \\
4\_Capture                     &  29 &  27 &   2 & CS0234(16), CS0115(11) \\
5\_Overcome                    &   1 &   1 &   0 & CS0234(1) \\
6\_Evade                       &  20 &   2 &  18 & CS0115(1), CS0234(1) \\
7\_Stealth                     &  14 &  12 &   2 & CS0234(7), CS0115(5) \\
8\_Herd\_Attract               &  22 &   4 &  18 & CS0115(2), CS0234(2) \\
9\_Conceal                     &  13 &   7 &   6 & CS0115(3), CS0239(2) \\
10\_Rescue                     &  14 &   5 &   9 & CS0246(5) \\
11\_Delivery                   &  27 &  27 &   0 & CS0234(16), CS0115(11) \\
12\_Guard                      &  16 &  16 &   0 & CS0234(8), CS0115(7) \\
13\_Race                       &   9 &   6 &   3 & CS0234(4), CS0115(1) \\
14\_Alignment                  &  21 &   5 &  16 & CS0234(3), CS0115(2) \\
15\_Configuration              &   3 &   1 &   2 & CS0246(1) \\
16\_Traverse                   &  21 &  21 &   0 & CS0234(19), CS0115(2) \\
17\_Survive                    &   8 &   7 &   1 & CS0115(4), CS0234(2) \\
18\_Connection                 &  18 &   6 &  12 & CS0115(2), CS0234(2) \\
19\_Exploration                &   3 &   0 &   3 & --- \\
20\_Reconnaissance             &  10 &   6 &   4 & CS0246(6) \\
21\_Contact                    &  21 &  19 &   2 & CS0234(9), CS0115(7) \\
22\_Enclosure                  &   5 &   3 &   2 & CS0115(1), CS0234(1) \\
23\_GainCompetence             &   3 &   2 &   1 & CS0246(1), CS0311(1) \\
24\_GainInformation            &  19 &  10 &   9 & CS0239(4), CS0115(3) \\
25\_LastManStanding            &  13 &   0 &  13 & --- \\
26\_KingoftheHill              &  20 &   4 &  16 & CS0246(2), CS0115(1) \\
\bottomrule
\end{tabularx}
\end{table}

\begin{table}[h]
\centering
\caption{Pattern-level errors: with\_schema\_free, DeepSeek-Coder-V2-Lite
(timeout 86--92\%; 20 logs per pattern).
G = grounding failure; H = hygiene failure}
\label{tab:pattern_free_ds}
\small
\begin{tabularx}{\linewidth}{@{}l r r r X@{}}
\toprule
Pattern & Total & G & H & Top 2 G codes \\
\midrule
1\_Ownership                   &  11 &   9 &   2 & CS0246(6), CS1061(3) \\
2\_Collection                  &  19 &   0 &  19 & --- \\
3\_Eliminate                   &  24 &   0 &  24 & --- \\
4\_Capture                     &  19 &   0 &  19 & --- \\
5\_Overcome                    &   8 &   0 &   8 & --- \\
6\_Evade                       &  14 &   0 &  14 & --- \\
7\_Stealth                     &  19 &   0 &  19 & --- \\
8\_Herd\_Attract               &  18 &   0 &  18 & --- \\
9\_Conceal                     &  20 &   0 &  20 & --- \\
10\_Rescue                     &  19 &   0 &  19 & --- \\
11\_Delivery                   &  14 &   0 &  14 & --- \\
12\_Guard                      &  17 &   0 &  17 & --- \\
13\_Race                       &   9 &   2 &   7 & CS0246(2) \\
14\_Alignment                  &  17 &   0 &  17 & --- \\
15\_Configuration              &  18 &   0 &  18 & --- \\
16\_Traverse                   &  16 &   1 &  15 & CS1061(1) \\
17\_Survive                    &  19 &   0 &  19 & --- \\
18\_Connection                 &  18 &   0 &  18 & --- \\
19\_Exploration                &  19 &   0 &  19 & --- \\
20\_Reconnaissance             &  17 &   0 &  17 & --- \\
21\_Contact                    &  14 &   0 &  14 & --- \\
22\_Enclosure                  &  20 &   0 &  20 & --- \\
23\_GainCompetence             &  19 &   0 &  19 & --- \\
24\_GainInformation            &  25 &   0 &  25 & --- \\
25\_LastManStanding            &  19 &   0 &  19 & --- \\
26\_KingoftheHill              &  18 &   0 &  18 & --- \\
\bottomrule
\end{tabularx}
\end{table}

\begin{table}[h]
\centering
\caption{Pattern-level errors: with\_schema\_free, Qwen2.5-Coder-7B
(timeout 86--92\%; 20 logs per pattern).
G = grounding failure; H = hygiene failure}
\label{tab:pattern_free_qwen}
\small
\begin{tabularx}{\linewidth}{@{}l r r r X@{}}
\toprule
Pattern & Total & G & H & Top 2 G codes \\
\midrule
1\_Ownership                   &  20 &   0 &  20 & --- \\
2\_Collection                  &  28 &   0 &  28 & --- \\
3\_Eliminate                   &  22 &   2 &  20 & CS0117(2) \\
4\_Capture                     &  28 &   0 &  28 & --- \\
5\_Overcome                    &  17 &   0 &  17 & --- \\
6\_Evade                       &  32 &   0 &  32 & --- \\
7\_Stealth                     &  22 &   0 &  22 & --- \\
8\_Herd\_Attract               &  25 &   1 &  24 & CS0246(1) \\
9\_Conceal                     &  27 &   0 &  27 & --- \\
10\_Rescue                     &  26 &   0 &  26 & --- \\
11\_Delivery                   &  49 &   0 &  49 & --- \\
12\_Guard                      &  24 &   0 &  24 & --- \\
13\_Race                       &  17 &   0 &  17 & --- \\
14\_Alignment                  &  20 &   0 &  20 & --- \\
15\_Configuration              &  27 &   0 &  27 & --- \\
16\_Traverse                   &  16 &   0 &  16 & --- \\
17\_Survive                    &  22 &   1 &  21 & CS1624(1) \\
18\_Connection                 &  19 &   0 &  19 & --- \\
19\_Exploration                &  42 &   0 &  42 & --- \\
20\_Reconnaissance             &  27 &   2 &  25 & CS0246(2) \\
21\_Contact                    &  39 &   0 &  39 & --- \\
22\_Enclosure                  &  33 &   0 &  33 & --- \\
23\_GainCompetence             &  25 &   0 &  25 & --- \\
24\_GainInformation            &  37 &   0 &  37 & --- \\
25\_LastManStanding            &  25 &   0 &  25 & --- \\
26\_KingoftheHill              &  17 &   0 &  17 & --- \\
\bottomrule
\end{tabularx}
\end{table}

\begin{table}[h]
\centering
\caption{Pattern-level errors: with\_schema\_min, DeepSeek-Coder-V2-Lite
(timeout 96\%; 20 logs per pattern).
G = grounding failure; H = hygiene failure}
\label{tab:pattern_min_ds}
\small
\begin{tabularx}{\linewidth}{@{}l r r r X@{}}
\toprule
Pattern & Total & G & H & Top 2 G codes \\
\midrule
1\_Ownership                   &  36 &  13 &  23 & CS0246(10), CS1061(3) \\
2\_Collection                  &  19 &   0 &  19 & --- \\
3\_Eliminate                   &  19 &   0 &  19 & --- \\
4\_Capture                     &  20 &   0 &  20 & --- \\
5\_Overcome                    &  23 &  11 &  12 & CS0246(6), CS1061(5) \\
6\_Evade                       &  20 &   4 &  16 & CS0246(2), CS1061(2) \\
7\_Stealth                     &  18 &   0 &  18 & --- \\
8\_Herd\_Attract               &  19 &   0 &  19 & --- \\
9\_Conceal                     &  20 &   0 &  20 & --- \\
10\_Rescue                     &  20 &   0 &  20 & --- \\
11\_Delivery                   &  20 &   0 &  20 & --- \\
12\_Guard                      &  18 &   0 &  18 & --- \\
13\_Race                       &  15 &   2 &  13 & CS0246(1), CS1061(1) \\
14\_Alignment                  &  24 &   0 &  24 & --- \\
15\_Configuration              &  19 &   0 &  19 & --- \\
16\_Traverse                   &  18 &   0 &  18 & --- \\
17\_Survive                    &  18 &   0 &  18 & --- \\
18\_Connection                 &  30 &   0 &  30 & --- \\
19\_Exploration                &  18 &   3 &  15 & CS0246(1), CS0311(1), CS1061(1) \\
20\_Reconnaissance             &  16 &   0 &  16 & --- \\
21\_Contact                    &  19 &   0 &  19 & --- \\
22\_Enclosure                  &  19 &   0 &  19 & --- \\
23\_GainCompetence             &  20 &   0 &  20 & --- \\
24\_GainInformation            &  26 &   4 &  22 & CS0246(4) \\
25\_LastManStanding            &  17 &   0 &  17 & --- \\
26\_KingoftheHill              &  18 &   0 &  18 & --- \\
\bottomrule
\end{tabularx}
\end{table}

\begin{table}[h]
\centering
\caption{Pattern-level errors: with\_schema\_min, Qwen2.5-Coder-7B
(timeout 96\%; 20 logs per pattern).
G = grounding failure; H = hygiene failure}
\label{tab:pattern_min_qwen}
\small
\begin{tabularx}{\linewidth}{@{}l r r r X@{}}
\toprule
Pattern & Total & G & H & Top 2 G codes \\
\midrule
1\_Ownership                   &  35 &   1 &  34 & CS0246(1) \\
2\_Collection                  &  31 &   0 &  31 & --- \\
3\_Eliminate                   &  24 &   7 &  17 & CS0246(4), CS8121(2) \\
4\_Capture                     &  18 &   0 &  18 & --- \\
5\_Overcome                    &  21 &   4 &  17 & CS0246(1), CS0311(1) \\
6\_Evade                       &  18 &   0 &  18 & --- \\
7\_Stealth                     &  18 &   0 &  18 & --- \\
8\_Herd\_Attract               &  17 &   1 &  16 & CS0246(1) \\
9\_Conceal                     &  20 &   0 &  20 & --- \\
10\_Rescue                     &  29 &   0 &  29 & --- \\
11\_Delivery                   &  24 &   4 &  20 & CS0246(4) \\
12\_Guard                      &  27 &   0 &  27 & --- \\
13\_Race                       &  21 &   2 &  19 & CS0246(2) \\
14\_Alignment                  &  23 &   0 &  23 & --- \\
15\_Configuration              &  26 &   9 &  17 & CS0246(5), CS0315(4) \\
16\_Traverse                   &  18 &   0 &  18 & --- \\
17\_Survive                    &  25 &   11 &  14 & CS0246(8), CS0311(2) \\
18\_Connection                 &  18 &   1 &  17 & CS0246(1) \\
19\_Exploration                &  20 &   0 &  20 & --- \\
20\_Reconnaissance             &  21 &   1 &  20 & CS1061(1) \\
21\_Contact                    &  18 &   0 &  18 & --- \\
22\_Enclosure                  &  27 &   2 &  25 & CS0246(2) \\
23\_GainCompetence             &  18 &   0 &  18 & --- \\
24\_GainInformation            &  17 &   0 &  17 & --- \\
25\_LastManStanding            &  20 &   2 &  18 & CS0246(2) \\
26\_KingoftheHill              &  20 &   0 &  20 & --- \\
\bottomrule
\end{tabularx}
\end{table}

\begin{table}[h]
\centering
\caption{Pattern-level errors: with\_schema\_full, DeepSeek-Coder-V2-Lite
(timeout 97--99\%; 20 logs per pattern).
G = grounding failure; H = hygiene failure}
\label{tab:pattern_full_ds}
\small
\begin{tabularx}{\linewidth}{@{}l r r r X@{}}
\toprule
Pattern & Total & G & H & Top 2 G codes \\
\midrule
1\_Ownership                   &  25 &  10 &  15 & CS0246(10) \\
2\_Collection                  &  19 &   0 &  19 & --- \\
3\_Eliminate                   &  22 &   0 &  22 & --- \\
4\_Capture                     &  17 &   0 &  17 & --- \\
5\_Overcome                    &  30 &   4 &  26 & CS0246(2), CS1061(2) \\
6\_Evade                       &  18 &   0 &  18 & --- \\
7\_Stealth                     &  18 &   0 &  18 & --- \\
8\_Herd\_Attract               &  17 &   0 &  17 & --- \\
9\_Conceal                     &  21 &   0 &  21 & --- \\
10\_Rescue                     &  20 &   0 &  20 & --- \\
11\_Delivery                   &  21 &   0 &  21 & --- \\
12\_Guard                      &  18 &   0 &  18 & --- \\
13\_Race                       &   8 &   3 &   5 & CS0246(3) \\
14\_Alignment                  &  20 &   1 &  19 & CS0246(1) \\
15\_Configuration              &  28 &   0 &  28 & --- \\
16\_Traverse                   &  22 &   0 &  22 & --- \\
17\_Survive                    &  20 &   0 &  20 & --- \\
18\_Connection                 &  19 &   0 &  19 & --- \\
19\_Exploration                &  20 &   0 &  20 & --- \\
20\_Reconnaissance             &  17 &   1 &  16 & CS0246(1) \\
21\_Contact                    &  16 &   0 &  16 & --- \\
22\_Enclosure                  &  20 &   0 &  20 & --- \\
23\_GainCompetence             &  23 &   1 &  22 & CS0246(1) \\
24\_GainInformation            &  22 &   0 &  22 & --- \\
25\_LastManStanding            &  18 &   0 &  18 & --- \\
26\_KingoftheHill              &  19 &   0 &  19 & --- \\
\bottomrule
\end{tabularx}
\end{table}

\begin{table}[h]
\centering
\caption{Pattern-level errors: with\_schema\_full, Qwen2.5-Coder-7B
(timeout 97--99\%; 20 logs per pattern).
G = grounding failure; H = hygiene failure}
\label{tab:pattern_full_qwen}
\small
\begin{tabularx}{\linewidth}{@{}l r r r X@{}}
\toprule
Pattern & Total & G & H & Top 2 G codes \\
\midrule
1\_Ownership                   &  24 &   6 &  18 & CS0246(5), CS0117(1) \\
2\_Collection                  &  28 &   8 &  20 & CS0246(6), CS1061(2) \\
3\_Eliminate                   &  21 &   0 &  21 & --- \\
4\_Capture                     &  18 &   0 &  18 & --- \\
5\_Overcome                    &  24 &   6 &  18 & CS0246(3), CS1061(3) \\
6\_Evade                       &  19 &   0 &  19 & --- \\
7\_Stealth                     &  26 &   6 &  20 & CS0246(3), CS1061(3) \\
8\_Herd\_Attract               &  19 &   0 &  19 & --- \\
9\_Conceal                     &  23 &   0 &  23 & --- \\
10\_Rescue                     &  26 &   0 &  26 & --- \\
11\_Delivery                   &  20 &   0 &  20 & --- \\
12\_Guard                      &  18 &   0 &  18 & --- \\
13\_Race                       &  19 &   1 &  18 & CS0246(1) \\
14\_Alignment                  &  35 &   9 &  26 & CS0246(4), CS1061(4) \\
15\_Configuration              &  28 &   0 &  28 & --- \\
16\_Traverse                   &  19 &   0 &  19 & --- \\
17\_Survive                    &  18 &   0 &  18 & --- \\
18\_Connection                 &  20 &   0 &  20 & --- \\
19\_Exploration                &  20 &   3 &  17 & CS0246(3) \\
20\_Reconnaissance             &  18 &   2 &  16 & CS0246(1), CS1061(1) \\
21\_Contact                    &  19 &   5 &  14 & CS0246(5) \\
22\_Enclosure                  &  26 &   1 &  25 & CS0246(1) \\
23\_GainCompetence             &  19 &   0 &  19 & --- \\
24\_GainInformation            &  19 &   0 &  19 & --- \\
25\_LastManStanding            &  17 &   0 &  17 & --- \\
26\_KingoftheHill              &  14 &   0 &  14 & --- \\
\bottomrule
\end{tabularx}
\end{table}

\end{document}